\documentclass[runningheads]{llncs}

\usepackage[T1]{fontenc}
\usepackage{authblk} 
\usepackage{xcolor}
\usepackage{algpseudocode}
\usepackage{amsmath}
\usepackage{amssymb}
\usepackage{amsfonts}
\usepackage{multirow}
\usepackage{bm}
\usepackage{graphicx,verbatim}
\usepackage{booktabs}
\usepackage{hyperref}
\usepackage[T1]{fontenc}
\usepackage{marvosym} 
\makeatletter
\newif\if@restonecol
\makeatother

\usepackage[linesnumbered,ruled,vlined]{algorithm2e}
\usepackage{algpseudocode}
\usepackage{amsmath}
\newcommand{\samethanks}[1][\value{footnote}]{\footnotemark[#1]}

\begin{document}
\title{SparseMamba-PCL: Scribble-Supervised Medical Image Segmentation via SAM-Guided Progressive Collaborative Learning }
%

\author{Luyi Qiu \thanks{Both authors contributed equally to the work.}, Tristan Till \samethanks, Xiaobao Guo, Adams Wai-Kin Kong (\Letter)}  
\authorrunning{Anonymized Author et al.}
\institute{College of Computing and Data Science, Nanyang Technological University, Singapore \\
\email{adamskong@ntu.edu.sg}}

\maketitle{}     
\begin{abstract}
Scribble annotations significantly reduce the cost and labor required for dense labeling in large medical datasets with complex anatomical structures. However, current scribble-supervised learning methods are limited in their ability to effectively propagate sparse annotation labels to dense segmentation masks and accurately segment object boundaries. To address these issues, we propose a Progressive Collaborative Learning framework that leverages novel algorithms and the Med-SAM foundation model to enhance information quality during training. (1) We enrich ground truth scribble segmentation labels through a new algorithm, propagating scribbles to estimate object boundaries. (2) We enhance feature representation by optimizing Med-SAM-guided training through the fusion of feature embeddings from Med-SAM and our proposed Sparse Mamba network. This enriched representation also facilitates the fine-tuning of the Med-SAM decoder with enriched scribbles. (3) For inference, we introduce a Sparse Mamba network, which is highly capable of capturing local and global dependencies by replacing the traditional sequential patch processing method with a skip-sampling procedure. Experiments on the ACDC, CHAOS, and MSCMRSeg datasets validate the effectiveness of our framework, outperforming nine state-of-the-art methods. Our code is available at \href{https://github.com/QLYCode/SparseMamba-PCL}{SparseMamba-PCL.git}.


\end{abstract}

\section{Introduction}

\label{sec:intro}
Medical image segmentation holds significant promise in computer-assisted diagnosis as it can identify regions of interest, such as organs, lesions, or tumors, which helps clinicians determine better treatment plans and follow-up strategies \cite{luo2022scribble,li2023scribblevc}. However, highly accurate segmentation requires large annotated datasets, which are costly and time-consuming to generate \cite{han2023scribble,li2024scribformer}. To address this challenge, researchers have investigated scribble-supervised methods \cite{chen2022scribble2d5,zhang2022shapepu,wang2023s} that rely on sparse annotations instead of pixel-to-pixel labels for model training. However, structural information is crucial for accurate segmentation, yet efforts to integrate it into learning frameworks have been largely neglected by previous work.

Learning from scribbles is a delicate process that requires a high-performance model and a sophisticated learning strategy. Recent advancements in high\linebreak-performance models for medical image segmentation rely heavily on Convolutional Neural Networks (CNNs) \cite{wang2020eca,han2022ghostnets,huang2021alignseg}, Transformers \cite{liu2021swin,cao2022swin,chen2024transunet}, and Mambas \cite{xing2024segmamba,liu2024swin1,chang2024net,wang2024lkm}. Mambas have gained considerable attention for their ability to capture global dependencies, addressing a limitation of CNNs, while maintaining linear computational complexity unlike the quadratic complexity of Transformers.
Mambas utilize Selective Scan 2D (SS2D) to process images from multiple directions, employing Selective State Space Models (S6) to handle each directional sequence for capturing global dependencies. However, multi-directional scanning may result in feature redundancy and masking of important patches, as each patch is scanned multiple times, making it challenging to determine the importance of each patch \cite{pei2024efficientvmamba}.

Using foundation models to train a segmentation model with weakly-\linebreak supervised data can greatly increase the prior knowledge of a training framework, as demonstrated by \cite{zhao2023segment}. Building on this, we guide the training of our model with Med-SAM, a variant of the Segment Anything Model (SAM), which has been fine-tuned on data from the medical domain \cite{chen2024ma}.
However, effectively utilizing Med-SAM for weakly-supervised training remains challenging: (1) Med-SAM’s performance heavily depends on input prompts, which are often suboptimal in weakly-supervised settings \cite{kirillov2023segment}; (2) static use of Med-SAM without fine-tuning during training limits both feature extraction capabilities of the encoder, as well as the segmentation performance of the decoder for expert tasks, and (3) Med-SAM struggles with segmentation precision along object boundaries.

To address these issues, we propose a novel scribble-supervised medical image segmentation framework via Progressive Collaborative Learning. The contributions of our framework are as follows:
\begin{enumerate} 
    \item Progressive Collaborative Learning (PCL) enhances collaborative learning to improve segmentation performance. It refines coarse segmentation masks into precise bounding box prompts, combines embeddings from two distinct encoders for richer feature representations, and iteratively fine-tunes the Med-SAM decoder to integrate expert knowledge during training.
    \item SparseMamba replaces the sequential image patch processing procedure in traditional Mambas with a modified skip-sampling algorithm to enhance the model’s capability for capturing global dependencies.
    \item Scribble-Propagated Object Boundary Estimator (SPOBE) leverages edge cues from images and scribbles to generate an auxiliary supervision signal, improving segmentation accuracy along object boundaries. 
    \item Extensive experiments are conducted on the ACDC, CHAOS, and MSCMR-\linebreak Seg datasets. Our framework outperforms 9 SOTA scribble-supervised segmentation methods.
\end{enumerate}

\section{The Proposed Framework}

\begin{figure}[htp] 
    \centering
    \includegraphics[width=0.95\linewidth]{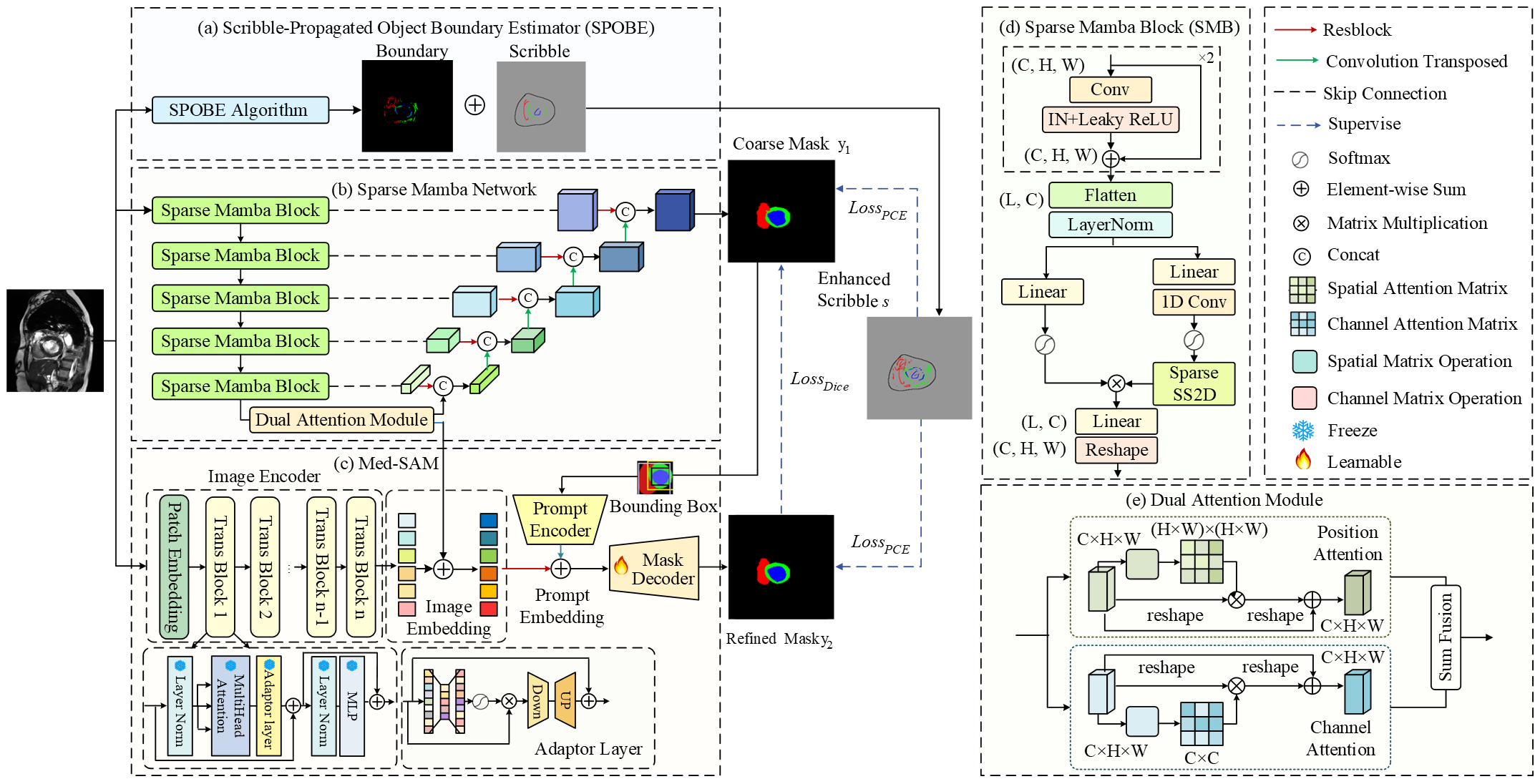}
    \caption{Overview of the SparseMamba-PCL framework for scribble-supervised image segmentation. (a) Scribble-Propagated Object Boundary Estimator, (b) SparseMamba, (c) Med-SAM Network, (d) Sparse Mamba Block, and (e) Dual Attention Module. \label{overall}} 
    \vspace{-7mm}
\end{figure}

SparseMamba-PCL is a weakly-supervised medical image segmentation framework that synergistically combines an object boundary estimator, a Med-SAM-guided training algorithm and a Sparse Mamba network (Fig.~\ref{overall}\footnote{Remark: Zoom in 400\% for better view.}). First, our framework extracts object boundary pixels as an auxiliary supervision signal and combines them with scribbles to generate enriched scribbles (Fig.~\ref{overall}(a)).
For training, we simultaneously process the input images using our Sparse Mamba network and Med-SAM (Fig.~\ref{overall}(b) and Fig.~\ref{overall}(c)). The Sparse Mamba network generates coarse segmentation masks, from which it extracts bounding boxes to prompt Med-SAM. Meanwhile, Med-SAM combines its encoder's image embeddings with those from the Sparse Mamba network, before merging them with the bounding box prompts. Finally, Med-SAM's mask decoder generates the refined segmentation masks. During training, these refined masks are used to optimize the Sparse Mamba network's weights. In addition, supervision is further enhanced by training both the Sparse Mamba network and Med-SAM with the enriched scribbles.

\subsection{Scribble-Propagated Object Boundary Estimator}
Reliably identifying boundaries of target objects is crucial for accurate segmentation. However, scribbles rarely provide quality boundary information, which makes weakly-supervised segmentation challenging. To alleviate the lack of boundary information, we propose a Scribble-Propagated Object Boundary Estimator (SPOBE), which exploits an edge detector and scribbles to identify target object boundaries in an iterative process. 

\begin{figure}[t] 
        \centering
    \includegraphics[width=0.9\linewidth]{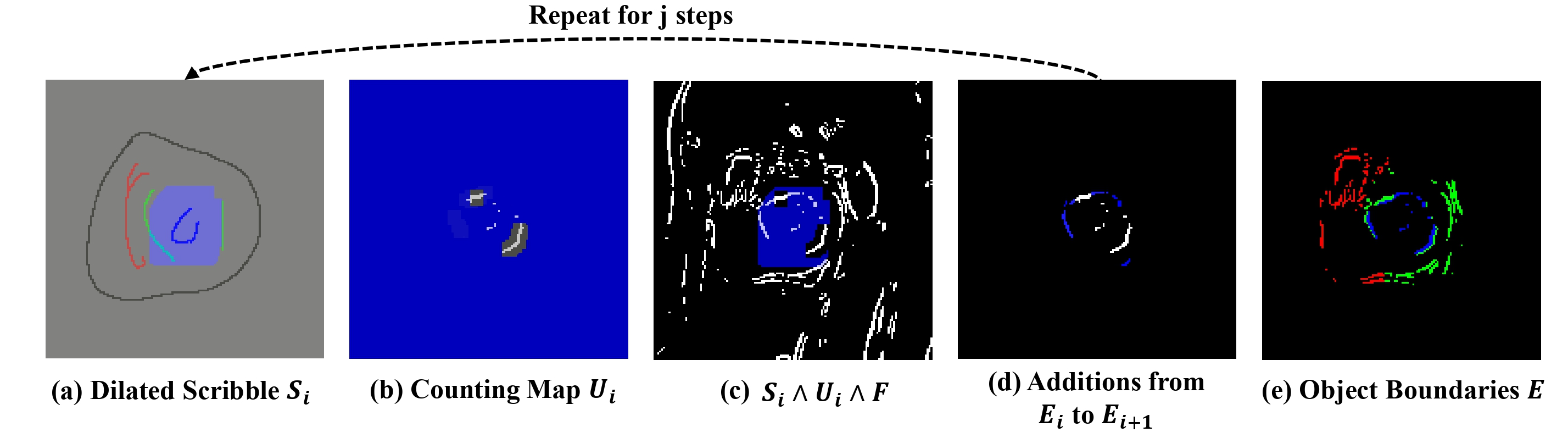}
    \vspace{-2mm}
    \caption{Visualization of one iteration in the boundary estimation algorithm. (a) $S_i$ shows the dilated scribble and (b) $U_i$ the counting map for the boundary map $E_i$ at the current step $i$. Blue regions in (a) and (b) represent pixels, for which the operations in $S_i$ and $U_i$ hold true respectively. $S$ and $U$ are logically combined with the edge map $F$ to extract edge pixels, which are added to update the boundary map to $E_{i+1}$. The process is repeated $j$ times for all kernel sizes and for every class to generate the final boundary map $E$, as shown in (e). \label{fig:obe} }
\end{figure}

First, an edge detector is applied to the input image and a full edge map $F$ is obtained. However, the edge map contains many edges unrelated to a given target object's boundary (Fig.~\ref{fig:obe}(c)). To distinguish the boundary edges from the noisy edges, we design an iterative scheme. Fig.~\ref{fig:obe} visualizes the procedure for a single iteration. In the first iteration, we initialize a counting map with same size as the input image and use a square kernel with a size of $k_1$ to dilate the scribbles (Fig.~\ref{fig:obe}(a) and (b)). The counting map and dilated scribbles for class $c$ are denoted as $U_{1,c}$ and $S_{1,c}$, respectively. Similar notations, $U_{i,c}$ and $S_{i,c}$ are used for the $i$-th iteration. The initial boundary edges are identified through $E_{1,c}=S_{1,c}\wedge U_{1,c}\wedge F$, as seen in Fig.~\ref{fig:obe}(c) and ~\ref{fig:obe}(d), where $\wedge$ is the logical AND operator. In the second iteration, a larger square kernel with a size of $k_2$ is used to dilate the scribbles and obtain $S_{2,c}$. The counting map value $U_{2,c}(x,y)=1$ if the total value of the $E_{1,c}$ patch centered at $(x,y)$ with a size of $k_2 \times k_2$ is less than $n_c$, which is a predefined threshold; otherwise $U_{2,c} (x,y)=0$. This threshold sets an upper limit on how many edge pixels can be added to the object boundary map at each iteration, reducing the risk of mistaking noisy edge-pixels with true boundary pixels as kernel sizes grow larger. The boundary edges identified in the second iteration are $E_{2,c}=S_{2,c}\wedge U_{2,c}\wedge F$. This process is repeated $j$ times with increasing kernel sizes. Fig~\ref{fig:obe}(e) shows an example of the final object boundaries. 

\subsection{Sparse Mamba Network} 
Fig.~\ref{overall}(b) shows SparseMamba, an encoder-decoder network with Sparse Mamba blocks in the encoder to capture local and global dependencies. The decoder, using residual blocks and transposed convolutions, preserves details and resolution. U-Net-style skip connections fuse hierarchical features between encoder and decoder. SparseMamba also incorporates a dual attention module \cite{fu2019dual} (Fig.~\ref{overall}(e)) to model spatial and channel dependencies. The decoder output is passed through a convolutional layer and Softmax activation to predict the segmentation probability map.

\textbf{Sparse Mamba Block (SMB).}
As shown in Fig.~\ref{overall}(d), input features of size \((C, H, W)\) pass through two residual blocks, then are flattened and transposed to \((L, C)\), where \(L = H \times W\). Features are processed in two parallel branches: the first expands them to \((2L, C)\) using a linear layer and a SiLU activation; the second applies a linear layer, 1D convolution, SiLU, and Sparse SS2D. The outputs are merged via the Hadamard product, projected back to \((L, C)\), and reshaped to \((C, H, W)\).

\begin{figure}[t] 
    \centering

    \includegraphics[width=\linewidth]{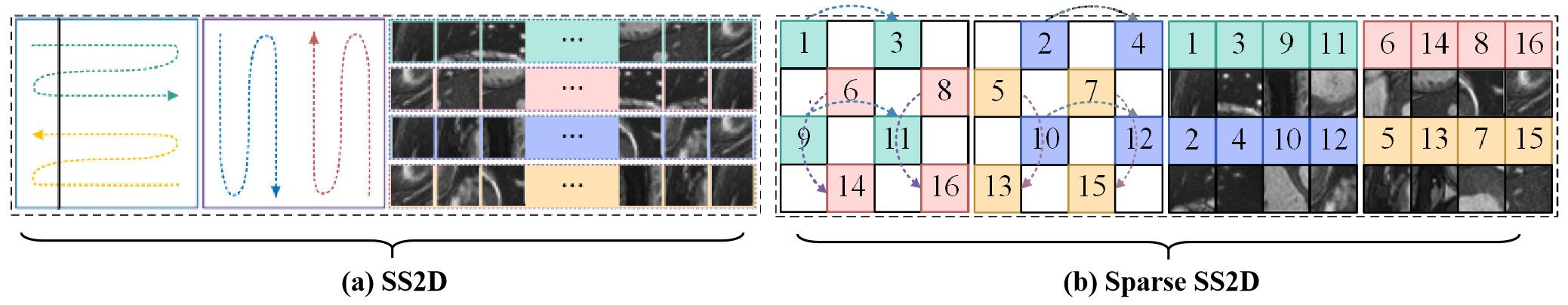}
    \vspace{-6mm}
    \caption{ (a) Vmamba \cite{liu2024vmamba} uses the 2D Selective Scan (SS2D) method, scanning four directions. (b) Sparse SS2D omits some sampling steps and performs intra-group traversal with a skipping step of 2. }
     \label{SparseSS2D}

\end{figure}

\textbf{Sparse SS2D.}
As Fig.~\ref{SparseSS2D}(a) demonstrates, SS2D scans each image from two starting positions (top-left and bottom-right) and two spatial orientations (vertically and horizontally), resulting in four independent scanning operations. Each directional sequence is processed by an S6 block to capture global dependencies.
However, multi-directional scanning introduces redundancy, as each patch is scanned multiple times, which may mask important patches and cause difficulties in determining their significance.
To address this, Sparse SS2D (Fig.~\ref{SparseSS2D}(b)) employs skip sampling \cite{pei2024efficientvmamba}, scanning each patch exactly once. This strategy reduces redundancy and enhances the representation of spatial relationships. 

\subsection{Progressive Collaborative Learning}
Progressive Collaborative Learning (PCL) is our proposed training framework, leveraging Med-SAM's anatomical priors to guide SparseMamba's prediction and refine the network's weights. 
First, as outlined in Algorithm \ref{alg:cap}, both encoders of SparseMamba and Med-SAM compute image embeddings independently. These embeddings are then summed to be a single fused embedding, which incorporates the spatial and structural information of both encoders for a richer and more diversified feature representation. Next, we use SparseMamba's outputs, our coarse segmentation mask $y_1$, to extract bounding box prompts for Med-SAM. Med-SAM’s performance heavily depends on accurate prompts, as vague prompts lead to suboptimal segmentation due to the structural complexity of medical images \cite{kirillov2023segment}. We combine SparseMamba's bounding boxes with bounding boxes extracted from scribbles to increase prompting stability in the early stages of training, in which $y_1$ is not trained sufficiently. 
These prompts are then processed by Med-SAM's prompt encoder to produce prompt embeddings. Finally, the fused image embeddings and prompt embeddings are merged into a single representation, which the Med-SAM decoder uses to generate a refined segmentation mask $y_2$ with improved accuracy and boundary precision.
We fine-tune SparseMamba and the Med-SAM decoder using these outputs. For SparseMamba, the difference between $y_1$ and $y_2$ is computed with Dice loss \small{$L_{\text{Dice}}(y_1, y_2) = 1 - \frac{2\,\langle y_1, y_2 \rangle}{\| y_1 \| + \| y_2 \|}$}. 
In addition, we calculate the partial cross-entropy loss \small{\( L_{pCE}(y, s) = - \sum_c \sum_{i \in \omega_s} \log (y_i^c) \)}, between the enriched scribble mask $E_S$ and $y_1$, as well as $E_S$ and $y_2$. Both terms are balanced by a weighting factor $\lambda=0.5$, resulting in a total loss $ L_{\text{total}}$ (Equation \ref{total}).
\begin{equation}
\label{total}
    L_{\text{total}}(y_1, y_2, E_S) = \lambda \times L_{\text{Dice}}(y_1, y_2) + (1-\lambda) \times \left( L_{\text{pCE}}(y_1, E_S) + L_{\text{pCE}}(y_2, E_S) \right),
\end{equation}
which is used to optimize SparseMamba. On the other hand, to ensure Med-SAM can acquire task-specific knowledge during training, we fine-tune the Med-SAM decoder using $L_{\text{pCE}}(y_2, E_S)$.

\vspace{-3mm}

{\tiny  
\setlength{\baselineskip}{0.8\baselineskip} 

\begin{algorithm}[htp]
\caption{\textbf{Progressive Collaborative Learning}}
\label{alg:cap}

\KwIn{\( X, S \)  
\quad  // \( \mathbf{X:} \) Medical image train set, \( \mathbf{S:} \) Scribbles.}
\KwOut {\( L_{\text{1}}, L_{\text{2}} \)  
\quad // \( L_\text{1} \): SparseMamba loss, \( L_\text{2} \) Med-SAM decoder loss. }

\For{\( t = 1 \) \textbf{to} epochs}{
    \For{\( b = 1 \) \textbf{to} batches}{       
        \textbf{Extract Enriched Embeddings:}\\
        \( E_S \gets \text{SPOBE}(X, S) \) // Enriched Scribbles from Boundary Estimator\; 
        
        \textbf{Image Embedding Fusion:}\\
        \( \mathbf{I_s} \gets \text{Sparse}_{\text{enc}}(\mathbf{x}) \) // Extract features from SparseMamba's encoder\;
        \( \mathbf{I_m} \gets \text{SAM}_{\text{enc}}(\mathbf{x}) \)  // Extract features from Med-SAM's encoder\;
        \( \mathbf{I} \gets \mathbf{I_s} + \mathbf{I_m} \)  // Fuse features as image embeddings\;
        
        \textbf{Coarse Segmentation:}\\
        \( \mathbf{y}_1 \gets \text{SparseMamba}(\mathbf{x}) \) // Generate coarse masks\;
        
        \textbf{Refined Segmentation:}\\
        \( \mathcal{C} \gets \Gamma(\mathbf{y}_1) \) \hfill // Extract contours\;
        \( \mathbf{B} \gets \beta(\mathcal{C}) \) \hfill // Compute bounding boxes\;
        \( \mathbf{P} \gets g_{\text{prompt}}(\mathbf{B}) \) \hfill // Encode prompt embeddings \;
        \( \mathbf{y}_2 \gets \text{SAM}_{\text{dec}}(\mathbf{I}, \mathbf{P}) \) \hfill // Generate refined masks using Med-SAM\;
        
        \textbf{Loss Optimization:}\\
        \( L_\text{1} \gets L_{\text{total}}(y_1, y_2, E_S) \) // Loss for SparseMamba\; 
        \( L_\text{2} \gets L_{\text{pCE}}(y_2, E_S) \) // Loss for Med-SAM decoder\;
    }
}

\end{algorithm}

} 
\vspace{-0.2in}
\vspace{-1mm}

\section{Experiments}

\subsection{Experimental Settings}
\textbf{Datasets and Evaluation Metrics.}
We evaluate our method on three public datasets:
(1) ACDC \cite{luo2022scribble}: Cine-MRI images from 100 patients, with manual annotations for the Right Ventricle (RV), Left Ventricle (LV), and Myocardium (MYO). The dataset is divided into 70 training, 15 validation, and 15 testing cases.
(2) MSCMRseg \cite{mscmr1}: Late Gadolinium Enhancement (LGE) MRI scans from 45 cardiomyopathy patients, annotated for the RV, LV, and MYO, are divided into 25 training, 5 validation, and 20 testing cases.
(3) CHAOS \cite{kavur2021chaos}: T1-weighted abdominal MR images from 20 subjects with liver, kidney, and spleen, split into 70\% training, 15\% validation, and 15\% testing cases.
For scribble annotations, the ACDC dataset uses manually created scribbles \cite{luo2022scribble}, while MSCMRSeg and CHAOS datasets use ITK-Snap to annotate 1-pixel-wide scribbles \cite{chen2022scribble2d5}. All results are based on 5-fold cross-validation, evaluated using the Dice coefficient and 95\% Hausdorff Distance (HD95). 

\textbf{Implementation Details.}
The images and annotations are resized to the same resolution of 256$\times$256 pixels. During training, each image is normalized to the range $[0,1]$ and augmented with random rotations, random flips, and random noise. We optimize our model using SGD with a weight decay of \(10^{-4}\), a momentum of 0.9, and a poly-learning rate schedule with a batch size of 16 for 90k iterations. During testing, we use SparseMamba's output for prediction. All experiments are implemented in PyTorch, trained on NVIDIA 2080Ti GPUs, under consistent experimental conditions.

\subsection{Experimental Analysis}

\begin{table}[t]
\centering
\caption{Comparison of SparseMamba with Transformer-, and Mamba-based Methods on three datasets. All models are trained using only \( L_{pCE}(y, s) \), where \( y \) denotes the segmentation mask predicted by the network and \( s \) denotes the scribbles.}

\label{tab1}
 \scriptsize
 \setlength\tabcolsep{0.5pt}
 \renewcommand\arraystretch{0.85}
\begin{tabular}{@{}lcccccccc@{}}
\toprule
 \multirow{2.5}{*}{Method}       &  \multirow{2.5}{*}{Published}    &  \multirow{2.5}{*}{Type}    & \multicolumn{2}{c}{ACDC}  & \multicolumn{2}{c}{CHAOS} & \multicolumn{2}{c}{MSCMRSeg} \\ \cmidrule(r){4-9}
             &              &         & Dice$_{(\%)}$ & HD95$_{(\text{mm} )}$ & Dice$_{(\%)}$  & HD95$_{(\text{mm} )}$ & Dice$_{(\%)}$   & HD95$_{(\text{mm} )}$   \\ \midrule


Swin Trans  \cite{liu2021swin}   & CVPR21       &   \multirow{3}{*}{Trans}      & 75.81       & \textbf{35.62}       & 57.15        & \underline{48.36}      & 65.95         & \textbf{64.57}        \\
Swin UNet  \cite{cao2022swin}    & ECCV22       &         & 75.58       & 49.78       & 54.73        & 53.04      & 65.12         & 84.87        \\
Trans UNet  \cite{chen2024transunet}    & MIA24        &         & 70.83       & 152.26      & 41.13        & 127.77     & 52.87         & 238.29       \\ \midrule 
SegMamba  \cite{xing2024segmamba}     & MICCAI24   &    \quad  \multirow{3}{*}{Mamba}        & \underline{79.86}       & 46.01       & \underline{63.03}        & 50.21      & \underline{68.16}         & 82.31        \\
LKM-UNet \cite{wang2024lkm}    & MICCAI24     &         & 77.41       & 48.80       & 60.12        & 53.67      & 66.24         & 87.21        \\
SwinUMamba  \cite{liu2024swin} \quad    & MICCAI24     &         & 71.19       & 92.30       & 51.77        & 87.93      & 55.93         & 165.38       \\
EM-Net \cite{chang2024net}     & MICCAI24     &         & 74.46       & 55.07       & 56.58        & 57.89      & 64.01         & 91.34        \\
 \midrule
SparseMamba & \multicolumn{2}{c}{Proposed} & \textbf{81.47}       & \underline{39.95}       & \textbf{68.01}        & \textbf{42.63}      & \textbf{73.25}         & \underline{69.77}        \\ \bottomrule
\end{tabular}
\end{table}

Table~\ref{tab1} compares SparseMamba with Transformer-, and Mamba-based methods across the three datasets. Transformers generally improve Dice scores but exhibit unstable HD95 results, achieving both the highest HD95 values (Trans UNet) on the three datasets and the lowest HD95 values (SwinTrans) on two datasets. In comparison, Mambas show consistently good performance for Dice and HD95. SparseMamba achieves the highest Dice on the three datasets, while producing the second lowest HD95 on ACDC and MSCMRSeg, as well as the lowest HD95 on the CHAOS dataset. These results demonstrate that SparseMamba offers an improvement in both segmentation accuracy and boundary precision compared to existing methods.

\begin{figure}[t]
    \centering 
   \setlength\tabcolsep{1pt}
    \begin{tabular}{cccccccccccc}

   \includegraphics[width=0.078\linewidth]{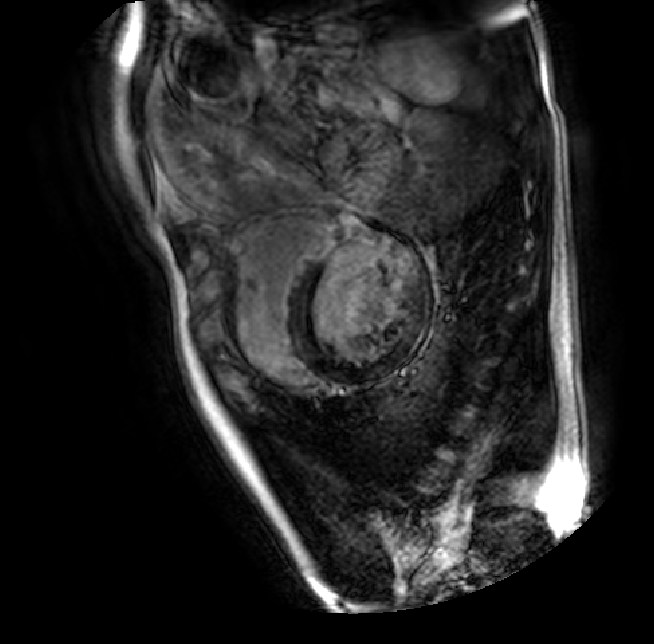}      &   \includegraphics[width=0.078\linewidth]{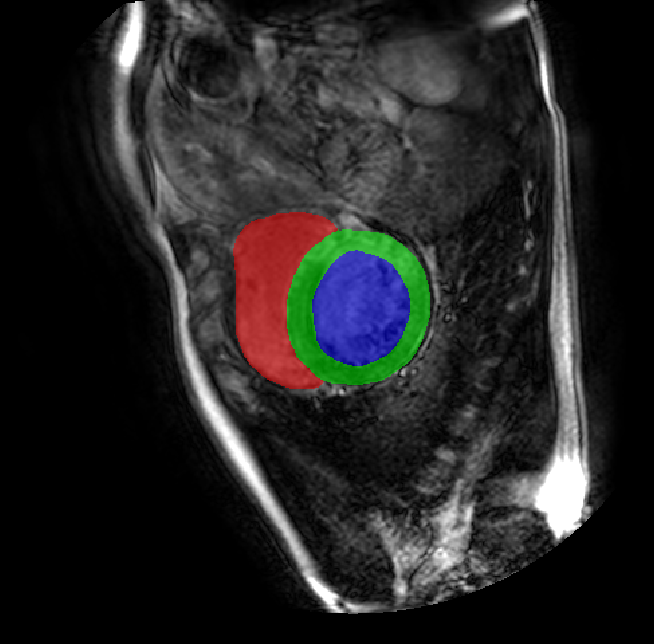} &    \includegraphics[width=0.078\linewidth]{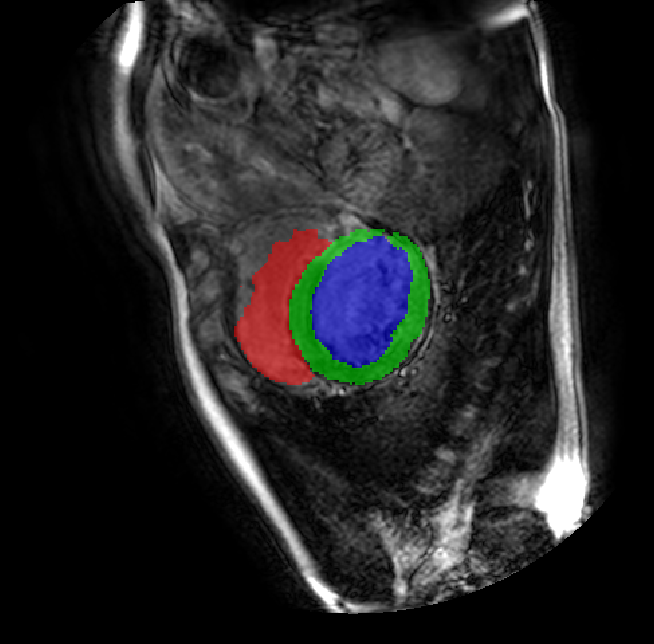}  &    \includegraphics[width=0.078\linewidth]{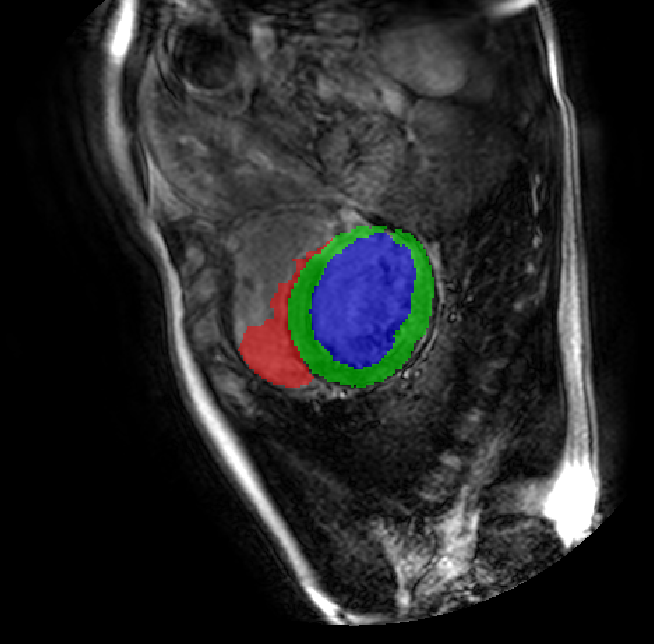} &   \includegraphics[width=0.078\linewidth]{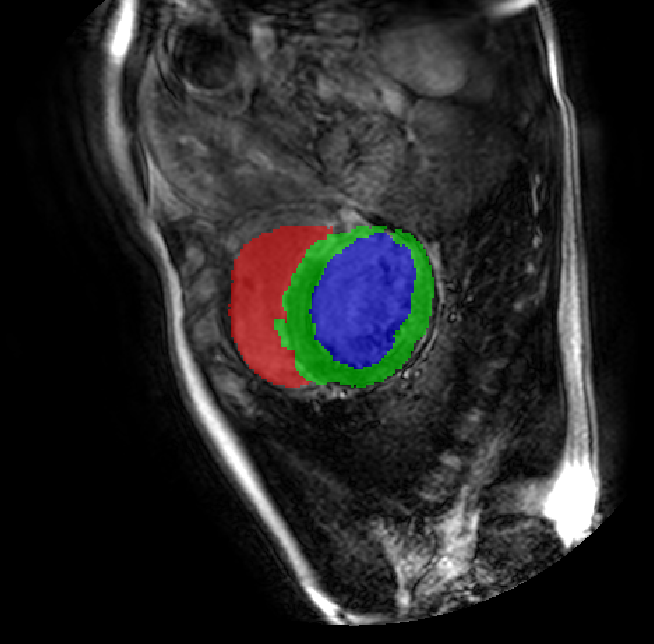} &   \includegraphics[width=0.078\linewidth]{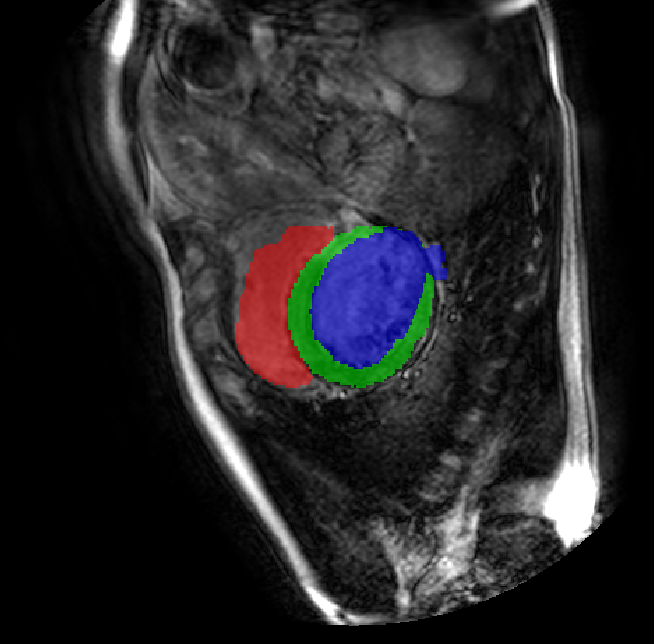} &   \includegraphics[width=0.078\linewidth]{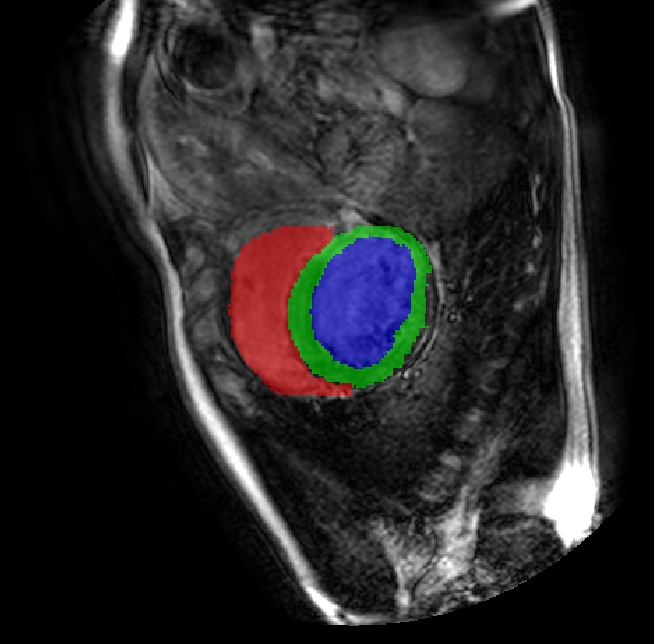} &   \includegraphics[width=0.078\linewidth]{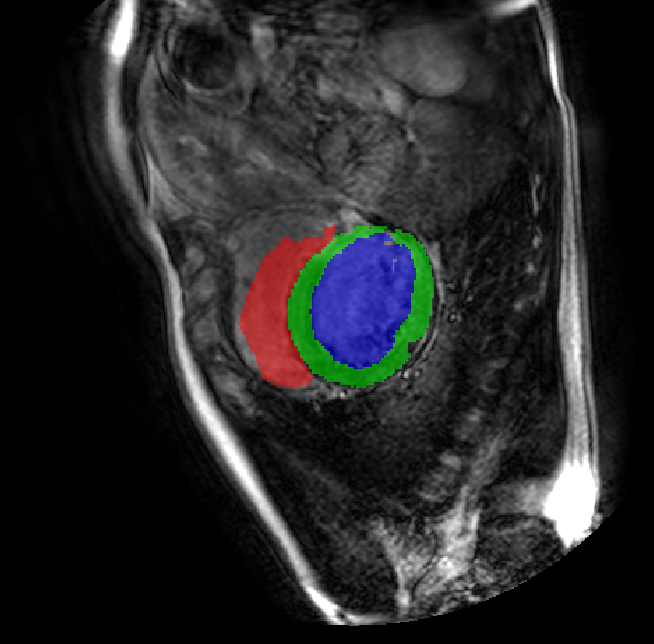} &   \includegraphics[width=0.078\linewidth]{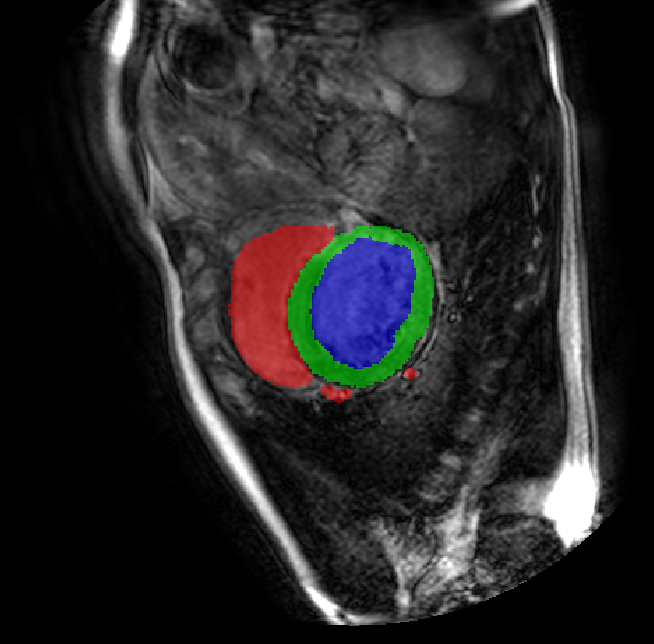} &   \includegraphics[width=0.078\linewidth]{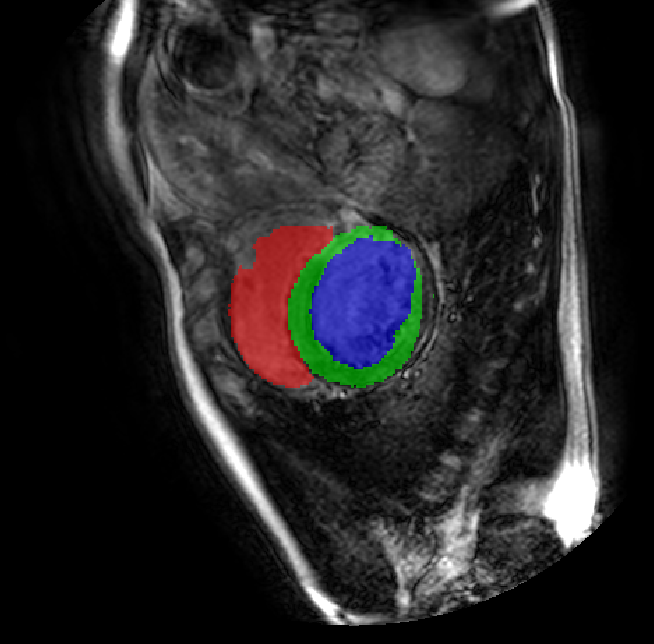} &   \includegraphics[width=0.078\linewidth]{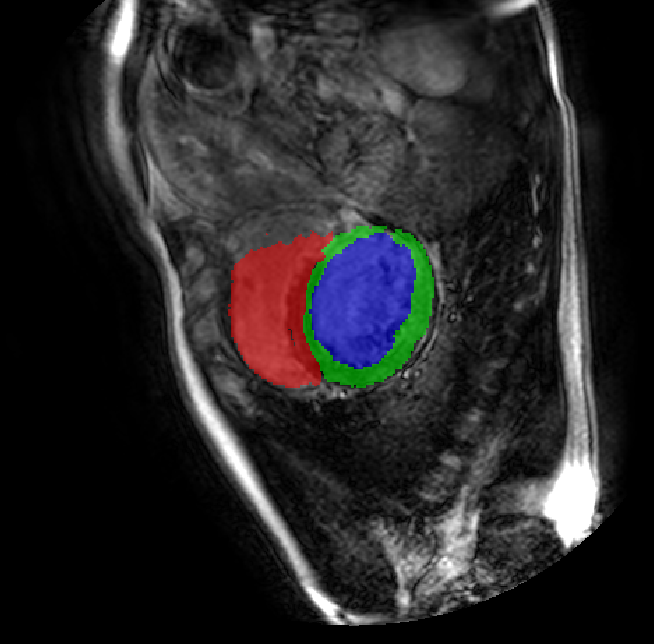} & 
   \includegraphics[width=0.078\linewidth]{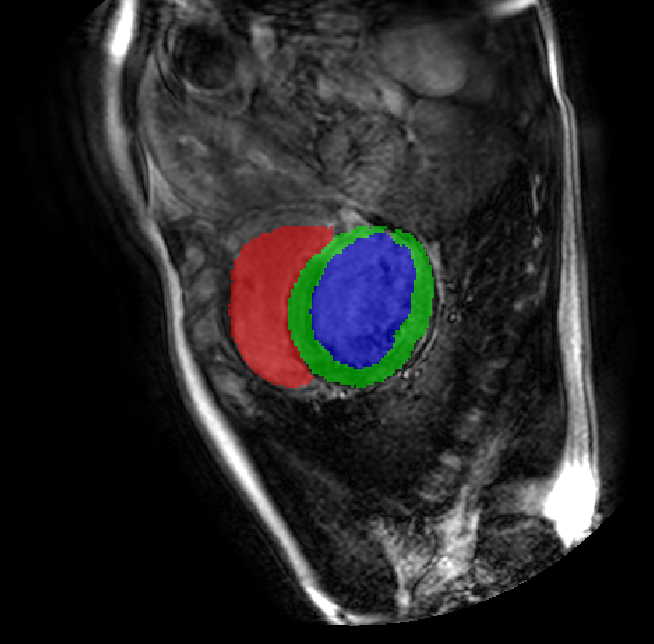}\\
   \includegraphics[width=0.078\linewidth]{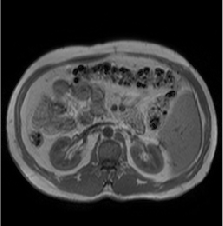} &   \includegraphics[width=0.078\linewidth]{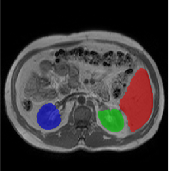} &    \includegraphics[width=0.078\linewidth]{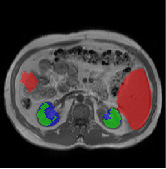} &    \includegraphics[width=0.078\linewidth]{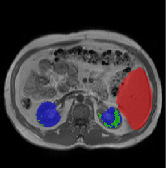} &   \includegraphics[width=0.078\linewidth]{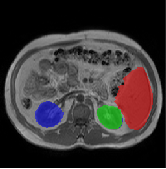} &   \includegraphics[width=0.078\linewidth]{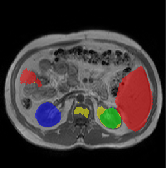} &   \includegraphics[width=0.078\linewidth]{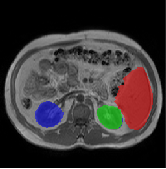} &   \includegraphics[width=0.078\linewidth]{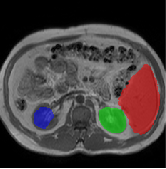} &   \includegraphics[width=0.078\linewidth]{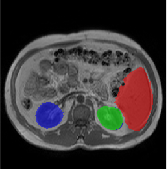} &   \includegraphics[width=0.078\linewidth]{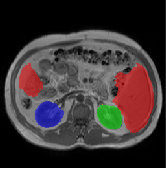} &   \includegraphics[width=0.078\linewidth]{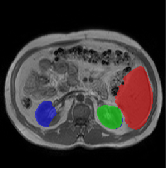} & 
   \includegraphics[width=0.078\linewidth]{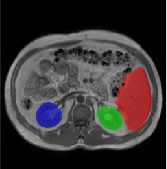}\\
   \includegraphics[width=0.078\linewidth]{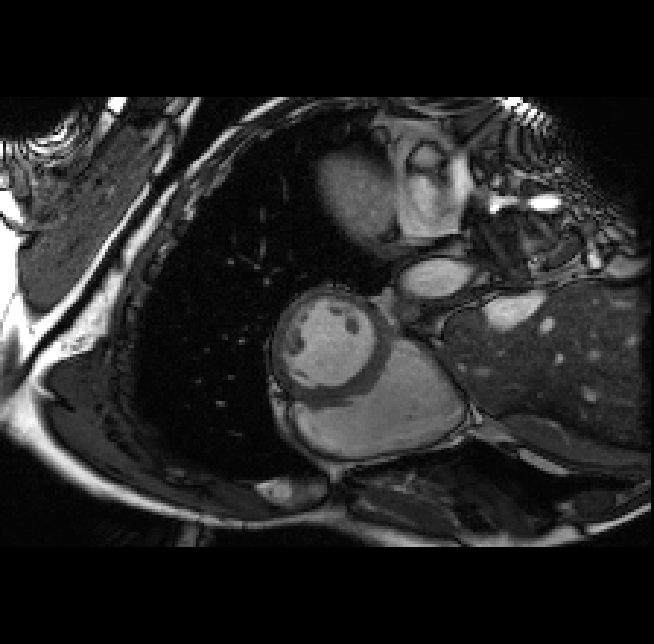}      &   \includegraphics[width=0.078\linewidth]{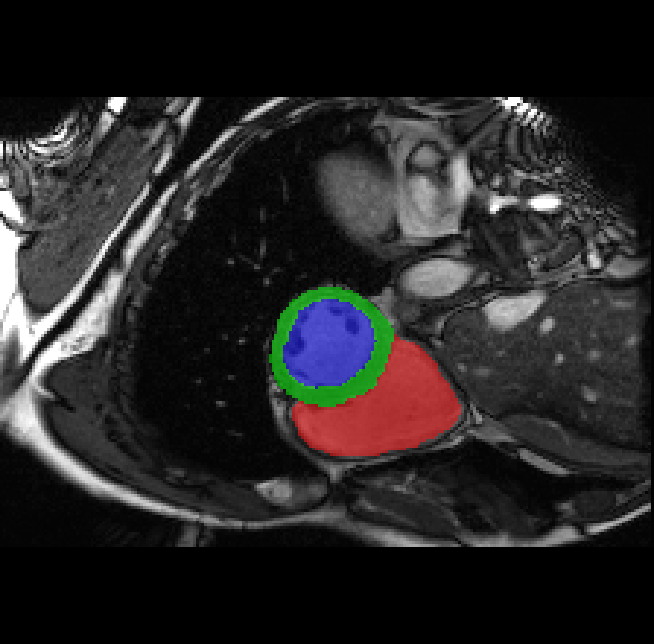} &    \includegraphics[width=0.078\linewidth]{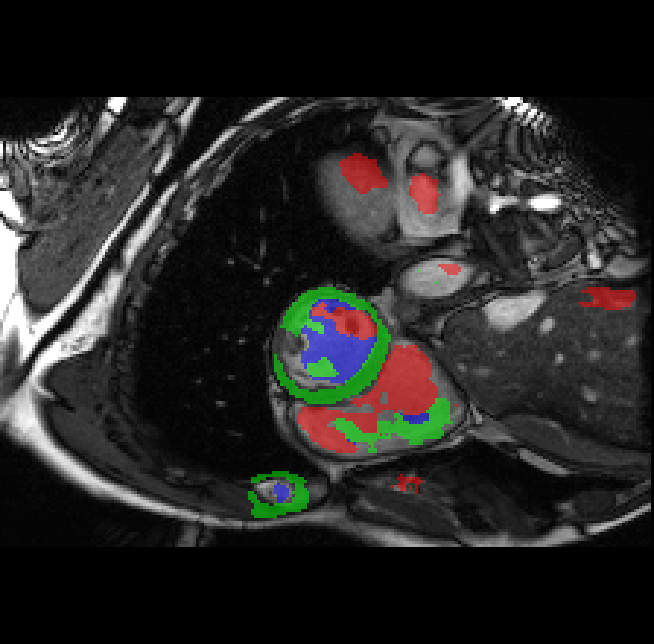}  &    \includegraphics[width=0.078\linewidth]{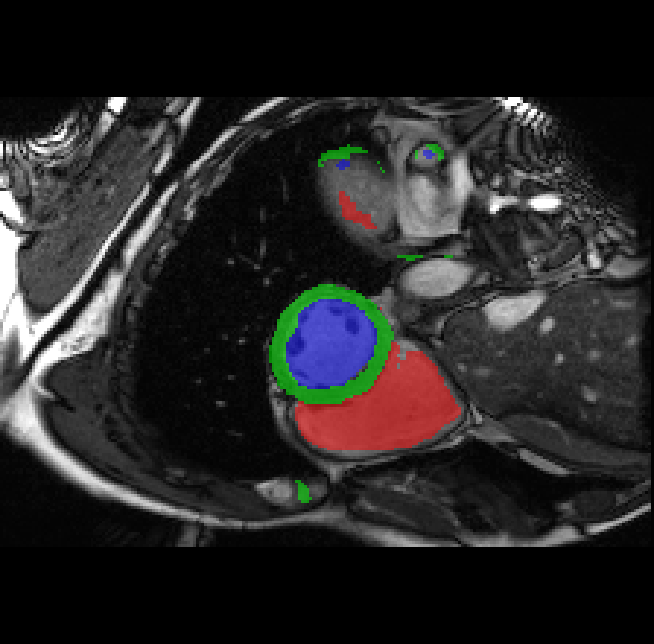} &   \includegraphics[width=0.078\linewidth]{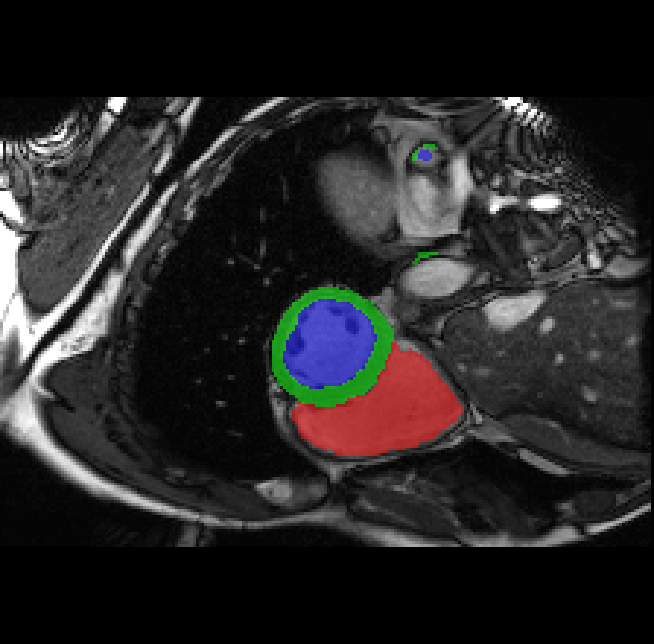} &   \includegraphics[width=0.078\linewidth]{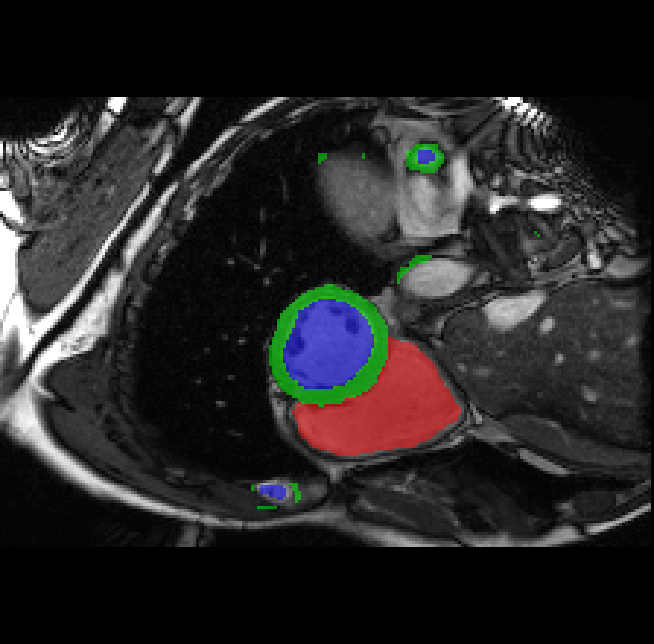} &   \includegraphics[width=0.078\linewidth]{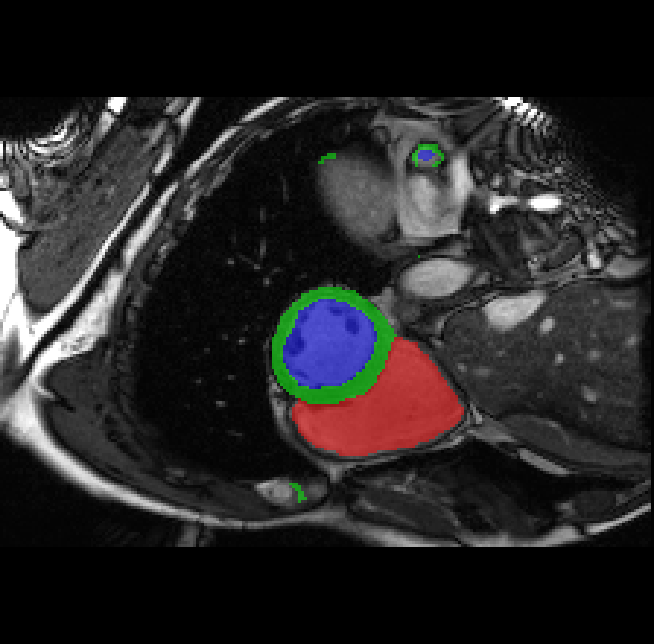} &   \includegraphics[width=0.078\linewidth]{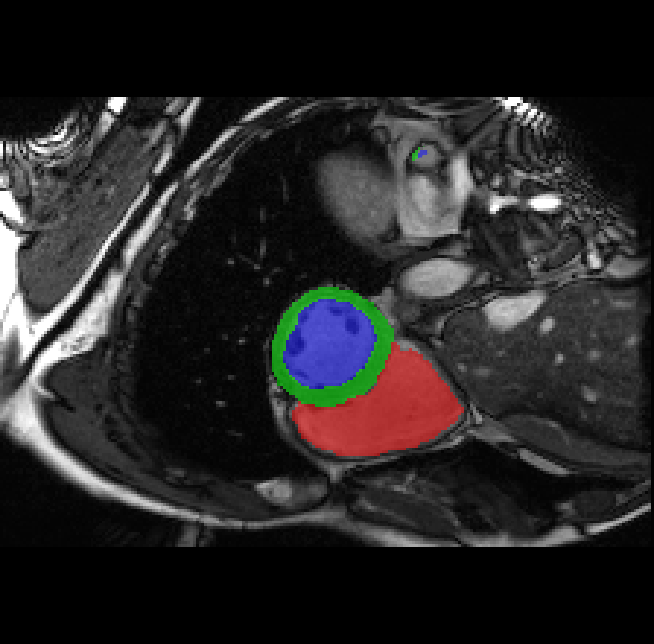} &   \includegraphics[width=0.078\linewidth]{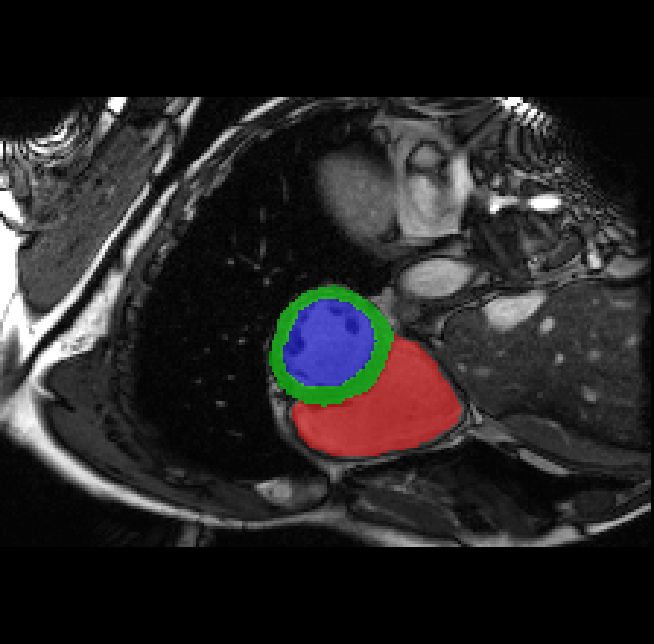} &   \includegraphics[width=0.078\linewidth]{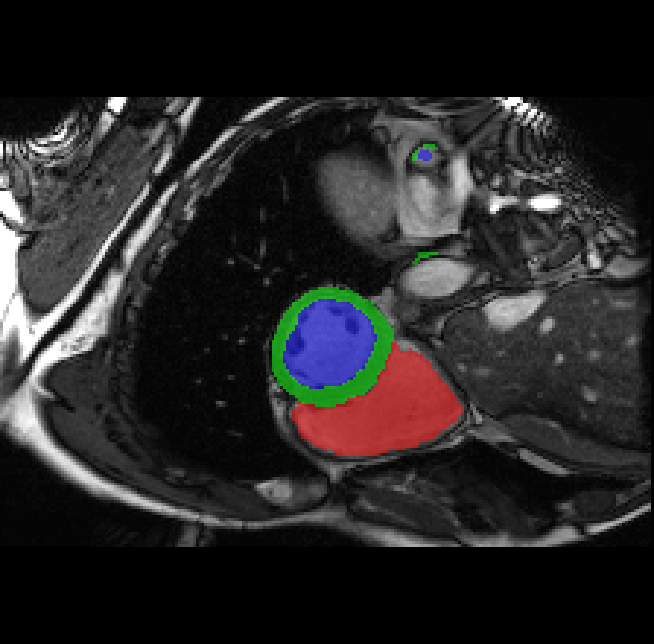} &   \includegraphics[width=0.078\linewidth]{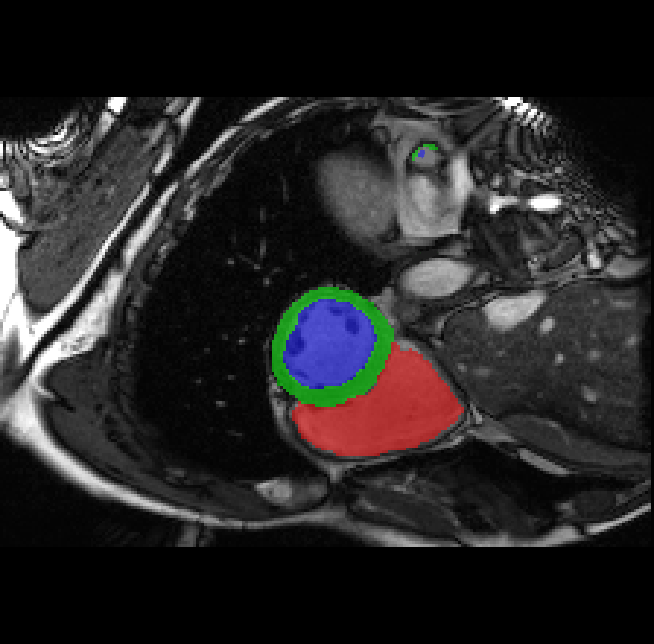} & 
   \includegraphics[width=0.078\linewidth]{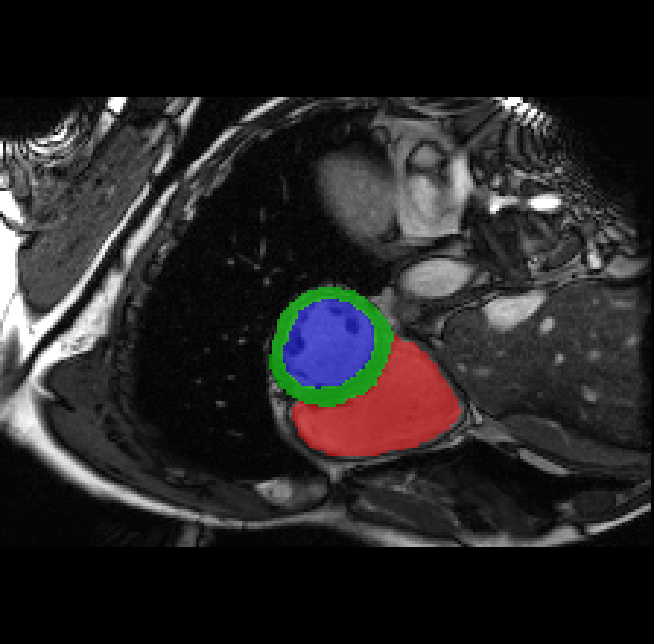}
   \\
   (a) & (b) & (c)& (d) & (e) & (f)&    (g) & (h) & (i)& (j) & (k) & (l) \\
    \end{tabular}

\caption{Qualitative comparison of weakly-supervised segmentation methods on ACDC, CHAOS, and MSCMRSeg datasets. (a) Input image, (b) ground truth, and segmentation results from (c) USTM \cite{liu2022weakly}, (d) Scribble2D5 \cite{chen2022scribble2d5}, (e) CycleMix \cite{zhang2022cyclemix}, (f) ShapePU \cite{zhang2022shapepu}, (g) S²ME \cite{wang2023s}, (h) ScribbleVC \cite{li2023scribblevc}, (i) TDNet \cite{han2023scribble}, (j) PacingPseudo \cite{yang2024non}, (k) Scribbleformer \cite{li2024scribformer}, and (l) SparseMamba-PCL are given.}   \label{visual}
\end{figure}

Table~\ref{tab2} presents an ablation study and compares SparseMamba-PCL with nine SOTA scribble-supervised methods across the three datasets. Out method achieves the highest Dice score across all the datasets, and  the lowest (CHAOS) and second lowest (ACDC and MSCMRSeg) HD95 values, confirming its effectiveness in segmentation accuracy and boundary refinement. Baseline+SPOBE and Baseline+PCL improve upon the Baseline (SparseMamba), demonstrating the benefits of boundary-aware supervision and SAM-guided learning. 
Fig.~\ref{visual} (l) compares the segmentation performance of SparseMamba-PCL with other scribble-supervised methods, showing smoother edges that precisely delineate object boundaries, unlike the jagged or blurred edges in other methods. The examples also demonstrate the consistent segmentation quality achieved by SparseMamba-PCL across ACDC, CHAOS, and MSCMRSeg, highlighting its adaptability across multiple medical domains. This adaptability and precise segmentation is crucial for accurate volumetric analysis and clinical decision-making, where even subtle boundary inaccuracies can lead to diagnostic errors. In summary, the SparseMamba-PCL architecture provides a consistent and robust improvement in segmentation metrics across diverse medical image datasets.

\begin{table}[t]
    \centering
    \caption{Comparison of scribble-supervised medical image segmentation methods on the three datasets. \textit{Note: \text{Baseline} corresponds to the SparseMamba results in Table~\ref{tab1}.}}\label{tab2}
    \scriptsize
     \renewcommand\arraystretch{0.85}
    \begin{tabular}{lccccccc}
    \toprule 
   \multirow{2}{*}{Method}& \multirow{2}{*}{Published} & \multicolumn{2}{c}{ACDC} & \multicolumn{2}{c}{CHAOS}&  \multicolumn{2}{c}{MSCMRSeg} \\ \cmidrule(r){3-8}
   & & Dice$_{(\%)}$ & HD95$_{(\text{mm})}$ & Dice$_{(\%)}$  & HD95$_{(\text{mm} )}$ & Dice$_{(\%)}$   & HD95$_{(\text{mm} )}$ \\ \midrule
USTM \cite{liu2022weakly}	& PR22	&78.81&	65.35	&48.12&	96.07&	51.73	&129.42\\ 
Scrbble2D5 \cite{chen2022scribble2d5}		&MICCAI22	&81.48	&33.59&	67.88	&43.65	&73.34	&67.85\\ 
 CycleMix\cite{zhang2022cyclemix}		&CVPR22	&86.57	&12.65	&71.19	&31.61	&77.46	  &57.53\\ 
ShapePU \cite{zhang2022shapepu}		&MICCAI22	&83.73	&22.37	&69.86	&41.59	&74.95	  &66.30\\ 
S$^2$ME \cite{wang2023s}		&MICCAI23	&86.44	&13.28	&70.67	&40.85	&76.59	 &63.21\\ 
ScibbleVC \cite{li2023scribblevc}		&ACMMM23&	88.13	&9.47	&72.31	&22.27	&80.84	&46.25\\
TDNet \cite{han2023scribble}		&MICCAI23	&88.46	&\textbf{5.03}&	\underline{73.02}&	\underline{17.93}	&\underline{81.96}&	\textbf{39.87}\\
PacingPseudo \cite{yang2024non}		&ESWA24	&85.92	&20.95	&70.15	&37.26	&75.86	&67.83\\ 
Scribbleformer \cite{li2024scribformer}		&TMI24	& \underline{88.63}	&5.85	&72.27	&19.04	&81.47 	&46.02\\\midrule

Baseline  &	 \multirow{4}{*}{Proposed} 	&81.47	&39.95	&68.01	&42.63	&73.25	&69.77 \\ 
Baseline+SPOBE&  		&84.29	&12.24	&70.03	&23.36	&75.88	&53.91\\ 
Baseline+PCL	&&87.82	&18.79	&71.65	&38.14	&78.92	&61.47\\ 
SparseMamba-PCL		& &\textbf{89.15}	&\underline{5.49}	&\textbf{73.84}	&\textbf{16.77}	&\textbf{82.13}	&\underline{41.79}\\ 
\bottomrule
    \end{tabular}
\end{table}

\section{Conclusion}

SparseMamba-PCL demonstrates a new approach to scribble-supervised segmentation by synergistically combining new algorithms and a medical foundation model to enhance boundary information and contextual understanding during training.  
This integrated approach significantly improves segmentation accuracy on three medical image datasets, outperforming nine SOTA methods. SparseMamba-PCL, with its novel boundary estimation algorithm, dynamic use of Med-SAM and efficient global dependency modeling, enables more robust and clinically relevant segmentation of complex anatomical structures.


\begin{thebibliography}{100}
\providecommand{\url}[1]{\texttt{#1}}
\providecommand{\urlprefix}{URL }
\providecommand{\doi}[1]{https://doi.org/#1}

\bibitem{swinunet}
Cao, H., et~al.: Swin-unet: Unet-like pure transformer for medical image segmentation. arXiv preprint arXiv:2105.05537  (2021)

\bibitem{chang2024net}
Chang, A., Zeng, J., Huang, R., Ni, D.: Em-net: Efficient channel and frequency learning with mamba for 3d medical image segmentation. In: International Conference on Medical Image Computing and Computer-Assisted Intervention. pp. 266--275. Springer (2024)

\bibitem{chen2024ma}
Chen, C., Miao, J., Wu, D., Zhong, A., Yan, Z., Kim, S., Hu, J., Liu, Z., Sun, L., Li, X., et~al.: Ma-sam: Modality-agnostic sam adaptation for 3d medical image segmentation. Medical Image Analysis  \textbf{98},  103310 (2024)

\bibitem{chen2024transunet}
Chen, J., Mei, J., Li, X., Lu, Y., Yu, Q., Wei, Q., Luo, X., Xie, Y., Adeli, E., Wang, Y., et~al.: Transunet: Rethinking the u-net architecture design for medical image segmentation through the lens of transformers. Medical Image Analysis  \textbf{97},  103280 (2024)

\bibitem{chen2022scribble2d5}
Chen, Q., Hong, Y.: Scribble2d5: Weakly-supervised volumetric image segmentation via scribble annotations. In: International Conference on Medical Image Computing and Computer-Assisted Intervention. pp. 234--243. Springer (2022)

\bibitem{fu2019dual}
Fu, J., Liu, J., Tian, H., Li, Y., Bao, Y., Fang, Z., Lu, H.: Dual attention network for scene segmentation. In: Proceedings of the IEEE/CVF conference on computer vision and pattern recognition. pp. 3146--3154 (2019)

\bibitem{han2022ghostnets}
Han, K., Wang, Y., Xu, C., Guo, J., Xu, C., Wu, E., Tian, Q.: Ghostnets on heterogeneous devices via cheap operations. International Journal of Computer Vision  \textbf{130}(4),  1050--1069 (2022)

\bibitem{huang2022scribble}
Huang, Z., Xiang, T.Z., Chen, H.X., Dai, H.: Scribble-based boundary-aware network for weakly supervised salient object detection in remote sensing images. ISPRS Journal of Photogrammetry and Remote Sensing  \textbf{191},  290--301 (2022)

\bibitem{huang2021alignseg}
Huang, Z., Wei, Y., Wang, X., Liu, W., Huang, T.S., Shi, H.: Alignseg: Feature-aligned segmentation networks. IEEE Transactions on Pattern Analysis and Machine Intelligence  \textbf{44}(1),  550--557 (2021)


\bibitem{kavur2021chaos}
Kavur, A.E., Gezer, N.S., Bar{\i}{\c{s}}, M., Aslan, S., Conze, P.H., Groza, V., Pham, D.D., Chatterjee, S., Ernst, P., {\"O}zkan, S., et~al.: Chaos challenge-combined (ct-mr) healthy abdominal organ segmentation. Medical Image Analysis  \textbf{69},  101950 (2021)

\bibitem{kirillov2023segment}
Kirillov, A., Mintun, E., Ravi, N., Mao, H., Rolland, C., Gustafson, L., Xiao, T., Whitehead, S., Berg, A.C., Lo, W.Y., et~al.: Segment anything. In: Proceedings of the IEEE/CVF International Conference on Computer Vision. pp. 4015--4026 (2023)

\bibitem{li2023scribblevc}
Li, Z., Zheng, Y., Luo, X., Shan, D., Hong, Q.: Scribblevc: Scribble-supervised medical image segmentation with vision-class embedding. In: Proceedings of the 31st ACM International Conference on Multimedia. pp. 3384--3393 (2023)

\bibitem{li2024scribformer}
Li, Z., Zheng, Y., Shan, D., Yang, S., Li, Q., Wang, B., Zhang, Y., Hong, Q., Shen, D.: Scribformer: Transformer makes cnn work better for scribble-based medical image segmentation. IEEE Transactions on Medical Imaging  \textbf{43}(6),  2254--2265 (2024)

\bibitem{liu2024swin}
Liu, J., Yang, H., Zhou, H.Y., Xi, Y., Yu, L., Li, C., Liang, Y., Shi, G., Yu, Y., Zhang, S., et~al.: Swin-umamba: Mamba-based unet with imagenet-based pretraining. In: International Conference on Medical Image Computing and Computer-Assisted Intervention. pp. 615--625. Springer (2024)

\bibitem{liu2024swin1}
Liu, J., Yang, H., Zhou, H.Y., Yu, L., Liang, Y., Yu, Y., Zhang, S., Zheng, H., Wang, S.: Swin-umamba†: Adapting mamba-based vision foundation models for medical image segmentation. IEEE Transactions on Medical Imaging  (2024)

\bibitem{liu2022weakly}
Liu, X., Yuan, Q., Gao, Y., He, K., Wang, S., Tang, X., Tang, J., Shen, D.: Weakly supervised segmentation of covid19 infection with scribble annotation on ct images. Pattern recognition  \textbf{122},  108341 (2022)

\bibitem{liu2024vmamba}
Liu, Y., Tian, Y., Zhao, Y., Yu, H., Xie, L., Wang, Y., Ye, Q., Liu, Y.: Vmamba: Visual state space model. arXiv preprint arXiv:2401.10166  (2024)

\bibitem{liu2021swin}
Liu, Z., Lin, Y., Cao, Y., Hu, H., Wei, Y., Zhang, Z., Lin, S., Guo, B.: Swin transformer: Hierarchical vision transformer using shifted windows. In: Proceedings of the IEEE/CVF international conference on computer vision. pp. 10012--10022 (2021)

\bibitem{luo2022scribble}
Luo, X., Hu, M., Liao, W., Zhai, S., Song, T., Wang, G., Zhang, S.: Scribble-supervised medical image segmentation via dual-branch network and dynamically mixed pseudo labels supervision. In: International Conference on Medical Image Computing and Computer-Assisted Intervention. pp. 528--538. Springer (2022)

\bibitem{pei2024efficientvmamba}
Pei, X., Huang, T., Xu, C.: Efficientvmamba: Atrous selective scan for light weight visual mamba. arXiv preprint arXiv:2403.09977  (2024)

\bibitem{wang2023s}
Wang, A., Xu, M., Zhang, Y., Islam, M., Ren, H.: S 2 me: spatial-spectral mutual teaching and ensemble learning for scribble-supervised polyp segmentation. In: International Conference on Medical Image Computing and Computer-Assisted Intervention. pp. 35--45. Springer (2023)

\bibitem{wang2024lkm}
Wang, J., Chen, J., Chen, D., Wu, J.: Lkm-unet: Large kernel vision mamba unet for medical image segmentation. In: International Conference on Medical Image Computing and Computer-Assisted Intervention. pp. 360--370. Springer (2024)

\bibitem{wang2020eca}
Wang, Q., Wu, B., Zhu, P., Li, P., Zuo, W., Hu, Q.: Eca-net: Efficient channel attention for deep convolutional neural networks. In: Proceedings of the IEEE/CVF conference on computer vision and pattern recognition. pp. 11534--11542 (2020)

\bibitem{xing2024segmamba}
Xing, Z., Ye, T., Yang, Y., Liu, G., Zhu, L.: Segmamba: Long-range sequential modeling mamba for 3d medical image segmentation. In: International Conference on Medical Image Computing and Computer-Assisted Intervention. pp. 578--588. Springer (2024)

\bibitem{yang2024non}
Yang, Z., Lin, D., Ni, D., Wang, Y.: Non-iterative scribble-supervised learning with pacing pseudo-masks for medical image segmentation. Expert Systems with Applications  \textbf{238},  122024 (2024)

\bibitem{zhang2022cyclemix}
Zhang, K., Zhuang, X.: Cyclemix: A holistic strategy for medical image segmentation from scribble supervision. In: Proceedings of the IEEE/CVF Conference on Computer Vision and Pattern Recognition. pp. 11656--11665 (2022)

\bibitem{zhang2022shapepu}
Zhang, K., Zhuang, X.: Shapepu: A new pu learning framework regularized by global consistency for scribble supervised cardiac segmentation. In: International Conference on Medical Image Computing and Computer-Assisted Intervention. pp. 162--172. Springer (2022)

\bibitem{zhao2023segment}
Zhao, Y., Zhou, T., Gu, Y., Zhou, Y., Zhang, Y., Wu, Y., Fu, H.: Segment anything model-guided collaborative learning network for scribble-supervised polyp segmentation. arXiv preprint arXiv:2312.00312  (2023)

\bibitem{mscmr1}
Zhuang, X.: Multivariate mixture model for myocardial segmentation combining multi-source images. IEEE TPAMI  \textbf{41}(12),  2933--2946 (2018)

\bibitem{han2023scribble}
Han, M., Luo, X., Liao, W., Zhang, S., Zhang, S., Wang, G.: Scribble-based 3d multiple abdominal organ segmentation via triple-branch multi-dilated network with pixel-and class-wise consistency. In: International Conference on Medical Image Computing and Computer-Assisted Intervention. pp. 33--42. Springer (2023)

\end{thebibliography}
\end{document}